\documentclass{article}
\usepackage{arxiv}
\usepackage{amsmath,graphicx}
\usepackage{xcolor}
\usepackage{plex-serif}
\usepackage[basic,italic,symbolgreek]{mathastext}
\usepackage{hyperref}

\def\L{{\cal L}}
\def\O{{\cal O}}
\def\N{{\cal N}}
\def\Tau{{\cal T}}
\usepackage{booktabs}
\title{Mitosis Detection Under Limited Annotation: A Joint Learning Approach}

\author{
 Pushpak Pati\\
 IBM Research Zurich,\\
 ETH Zurich\\
  \texttt{pus@zurich.ibm.com} \\
   \And
 Antonio Foncubierta-Rodriguez\\
 IBM Research Zurich\\
  \texttt{fra@zurich.ibm.com} \\
  \And
 Orcun Goksel\\
 ETH Zurich\\
  \texttt{ogoksel@vision.ee.ethz.ch} \\
  \And
 Maria Gabrani\\
 IBM Research Zurich\\
  \texttt{mga@zurich.ibm.com} \\
  }

\begin{document}
\maketitle
\copyrightnote{\copyright 2020 IEEE. Personal use of this material is permitted.
Permission from IEEE must be obtained for all other uses, in any current or future  media, including reprinting/republishing this material for advertising or promotional purposes, creating new collective works, for resale or redistribution to servers or lists, or reuse of any copyrighted component of this work in other works 
DOI: \href{https://doi.org/10.1109/ISBI45749.2020.9098431}{10.1109/ISBI45749.2020.9098431}
}
\begin{abstract}
Mitotic counting is a vital prognostic marker of tumor proliferation in breast cancer. Deep learning-based mitotic detection is on par with pathologists, but it requires large labeled data for training. We propose a deep classification framework for enhancing mitosis detection by leveraging class label information, via softmax loss, and spatial distribution information among samples, via distance metric learning. We also investigate strategies towards steadily providing informative samples to boost the learning. The efficacy of the proposed framework is established through evaluation on ICPR 2012 and AMIDA 2013 mitotic data. Our framework significantly improves the detection with small training data and achieves on par or superior performance compared to state-of-the-art methods for using the entire training data. 

\end{abstract}

\section{Introduction}
\label{sec:introduction}

Microscopic analysis of tissue samples, and in particular, mitotic cell count is a crucial and well-established indicator of tumor aggressiveness in breast cancer~\cite{elston91}.
In clinical practice, manual detection of mitosis in H\&E images is tedious, time-consuming and subject to inter-observer variability.
Rapid digitization of glass slides, along with the advances in deep learning strategies has empowered the development of image analysis algorithms for automated diagnosis, prognosis, and prediction in digital pathology~\cite{litjens17}.
Deep learning methods have successfully addressed several tasks in digital pathology by reducing the laborious and tedious nature of providing accurate quantification and by reducing the amount of inter-reader variability among pathologists, at the cost of requiring large amounts of task-specific, good quality annotated data~\cite{tizhoosh18}. 
Additionally, the high variability in the appearance of mitosis due to aberrant chromosomal makeup of many tumors and imperfections of the tissue preparation strongly underline the need for rich, large training datasets. 

Several efforts have been made to improve the mitosis detection under the availability of large annotated data by posing it as a classification task~\cite{ciresan13, cli18} or a semantic segmentation task~\cite{cli19}. 
However, no effort has been dedicated to tackling the issue of limited training data. 
Recent advancements in meta-learning and metric learning~\cite{kaya19} have paved the way for addressing the data scarcity issue~\cite{finn19}. 

In this work, we investigate deep metric learning and propose a classification method for mitosis detection utilizing both class label information and local spatial distribution information between training samples, to be able to learn from fewer annotations. The framework leverages the complementary information to enhance classification performance and learn better discriminative embeddings. 
We evaluate our framework on two popular mitotic datasets, ICPR 2012~\cite{icpr12} and AMIDA 2013~\cite{amida13} indicating significant improvement over pure classification based detection methods when using fewer annotated data. 
Our framework achieves state-of-the-art (SOTA) results for classification approaches and comparable results to semantic segmentation approaches upon using the complete labeled training data.

\section{Related work}
\label{sec:relatedwork}
In the past few years, a number of mitosis detection contests, including ICPR 2012 (ICPR12)~\cite{icpr12} and AMIDA 2013 (AMIDA13)~\cite{amida13} have not only pointed the relevance of the task, but also promoted remarkable advances in the area.
Methods generally follow a two-step procedure, with a coarse method for candidate mitotis localization, and a classifier for fine-grained decisions. IDSIA~\cite{ciresan13} proposes a computationally expensive sliding-window-based detection method. DeepMitosis~\cite{cli18} applies a proposal-based deep detection network for mitotic detection and a patch-based deep verification network to improve the predictions. SegMitos~\cite{cli19}  proposes a concentric loss based semantic segmentation approach to identify the mitoses. The detection performance improves with increasing the labeled training data~\cite{akram18}, but no work has focused on tackling the lack of sufficiently many annotations issue.

Softmax loss-based network training does not explicitly encourage intra-class compactness and inter-class separation. A recent study~\cite{horiguchi17} compares the softmax loss to the deep metric learning loss and indicates their complementary information. Therefore, a joint learning framework combining the classification and similarity constraints can improve performance.
Examples are combination of softmax loss with  triplet loss~\cite{zhang16} and pair loss~\cite{yli17}.
Several sampling strategies~\cite{facenet15,bh17,dws17} have been proposed to select informative samples to boost the deep metric learning competence. In this work, we propose a learning strategy that jointly optimizes softmax and triplet loss, and propose a triplet sampling strategy for steadily selecting informative samples improving mitotic detection.

\section{Materials and methods}
\label{sec:dataset}
We use data from ICPR12 and AMIDA13 mitotic detection contests.
H\&E stained images were acquired by Aperio XT scanner at 40X magnification for both datasets. Specific details of the datasets are given in Table~\ref{table:datasets}. ICPR12 contains per-pixel annotation and AMIDA13 contains point annotations for mitoses. We select the centroids of the labeled mitoses in ICPR12 to convert them into point annotations to develop a consistent framework for both the datasets.

\begin{table}[t]
\caption{ICPR12 and AMIDA13 datasets.}
\label{table:datasets}
\centering
\begin{tabular}{lcc}
\hline
\ & ICPR12 & AMIDA13 \\ 
\hline
High-Power Field (HPF) resolution & 0.2456 $\mu{m/px}$ & 0.25 $\mu{m/px}$ \\
Training HPFs & 35 & 311 \\
Test HPFs & 15 & 295 \\
Training mitoses & 226 & 550 \\ 
Test mitoses & 101 & 533 \\ \hline
\end{tabular}
\end{table}

We begin with normalizing the H\&E stained images using color deconvolution~\cite{stanisavljevic18} to reduce the variability due to tissue preparation and imaging techniques. Since in H\&E staining, nuclei possess higher blue channel intensity, we convert RGB images into blue-ratio images~\cite{chang12} and identify potential nuclei locations by detecting high brightness objects in them.
Morphological opening  followed by Otsu thresholding and connected components analysis result in potential nuclei locations, for connected components above 100 pixels.

\begin{figure}[t]
\centering
\centerline{\includegraphics[width=0.65\linewidth]{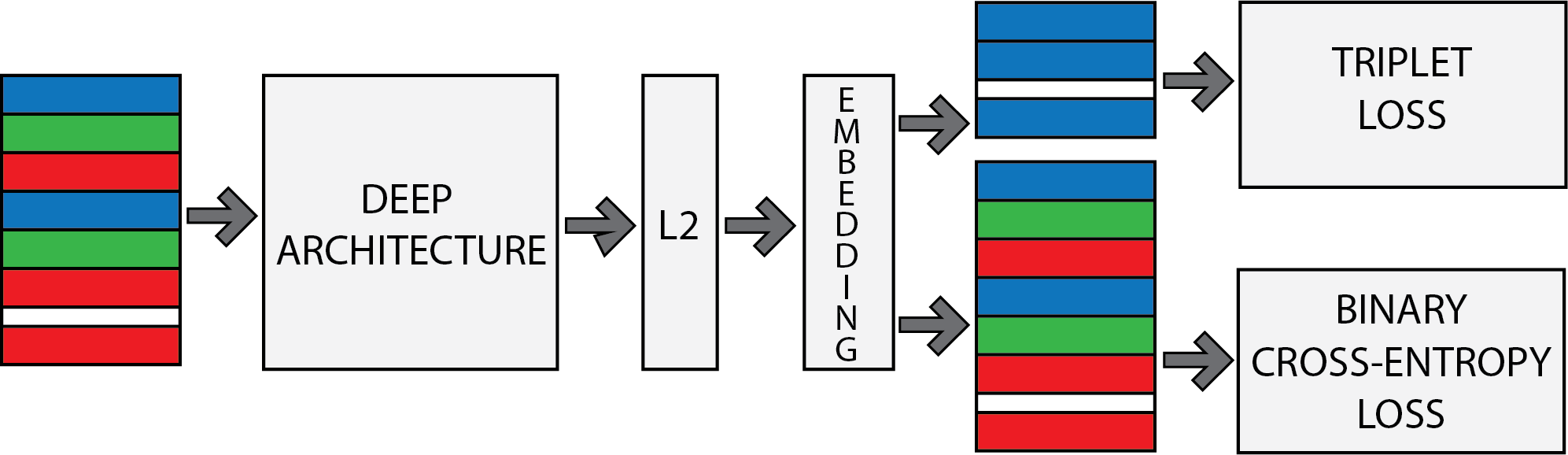}}
\caption{Framework structure, and color-coded samples: \textit{anchor} (blue), \textit{positive} (green) and \textit{negative} (red).}
\label{fig:network}
\end{figure}

\subsection{Network architecture, Loss function and Sampling}
We extract  patches $x$, of size $72\times72$, centered around the detected nuclei and employ our deep learning framework presented in Figure \ref{fig:network} to identify mitotic figures among the detected nuclei. We use Wide Residual Network (WRN)~\cite{zagoruyko16} as the deep architecture, $f_\theta$ parameterized by $\theta$, following its success in mitosis detection~\cite{erwan17}. Feature representation, $f(x)$, is obtained from WRN's penultimate layer and $f(x)$ is normalized to have unit length for training stability. We extract embeddings for $x$ and jointly optimize binary cross entropy (BCE) loss and triplet loss~\cite{facenet15}. 

A triplet consists of an anchor sample $a$, a positive one from the same class $p$ and a negative one $n$ from the opposite class. Triplet loss minimizes the distance $d_{a,p} = \|f(x_a)-f(x_p)\|^2$ and maximizes the distance $d_{a,n} =\|f(x_a)-f(x_n)\|^2$. With $\Tau$, the set of triplets with cardinality $T$, the joint loss $\L^{joint}(\Tau)$ is defined as,
\vspace{-2mm}
\begin{gather*}
    \L^{triplet}(\Tau) = \dfrac{1}{T}\sum_{i=1}^{T}\,\left[d_{a_i,p_i}-d_{a_i,n_i}+m\right]_+ \\
    \L_{joint}(\Tau) = \L_{bce}(\Tau_a) + \alpha \, \L_{triplet}(\Tau_{apn}) + \lambda \, ||W||^2
\end{gather*}

where $m$ denotes the enforced margin between $a,p$ and $a,n$ pairs. $\L^{joint}$ is determined by $\L^{bce}$, BCE loss for all anchors in $\Tau$, $\L^{triplet}$, triplet loss for all triplets in $\Tau$, and $L2$ regularization loss of network weights $W$. $\lambda$ is the coefficient for regularization and $\alpha>0$ is one trade-off parameter. In the training phase, $\L^{joint}$ is backpropagated for for updating the network, and in the testing phase, the classification result is generated with $\L^{bce}$.

With the number of training samples $N$, computing $\L^{triplet}$ over $\O(N^3)$ triplets is 
computationally infeasible. 
Since not all triplets do contribute to $\L^{triplet}$, as very few negatives violate the margin constraint, it is important to select informative triplets during the training phase. In a mini-batch $B$, the triplets are constructed in an online fashion~\cite{facenet15}, where each sample is set as the $a$ once, and the $p$ and the $n$ are selected in reference to $a$ of the triplet. To this end, we examine three triplet sampling strategies:

\textbf{Random sampling (RS)}:
Each sample has the same probability of being chosen, independently of $a$.

\textbf{Batch-hard sampling (BHS)}:
The hardest positive $p^*$ (furthest from $a$) and the hardest negative $n^*$ (closest to $a$) within $B$ are chosen for every $a$. 

\textbf{Distance weighted sampling (DWS)}:
The distribution of pairwise distances on a $D$-dimensional unit sphere, $q(d) \propto d^{n-2} \left[1 - \frac{d^2}{4}\right]^{\frac{n-3}{2}} \sim \N(\sqrt{2}, \frac{1}{2D})$ when $D>128$. 
Hence, we sample $n$ from the set of $d_{an}<\sqrt{2}$, as the complementary set induces no loss, and thus no progress for learning. 
For a given $a$ and a cut-off parameter $\beta$, DW sampling selects $p^*$ and $n^*$ following 
\vspace{-2mm}
\begin{gather*}
    Pr(p^* = p |a) \propto q(d_{ap}) \\
    Pr(n^* = n |a) \propto \min(\beta, q^{-1}(d_{an}))
\end{gather*}

\subsection{Training and testing scheme}
Most of the detected nuclei during pre-processing are non-mitoses that can be easily separated from the mitoses. A preliminary network is trained using all the mitoses and a subset of non-mitoses $NM$. The network identifies hard non-mitotic figures, non-mitoses predicted as mitoses with high probability, from the complementary set $NM'$. The hard non-mitoses are added to $NM$ for subsequent network training. $NM$ is periodically extended by repeating the above scheme. This scheme prevents fitting suboptimal hypotheses by using evidence from the entire training set and it significantly limits the non-mitotic samples in the training phase. During the testing phase, mitotic class probabilities are predicted for a number of neighboring patches around every nuclei. For a nucleus, the mean of its neighboring patch probabilities is thresholded to decide its class label.

\section{Experiments and results}
\label{sec:results}
We evaluate our method on ICPR12 and AMIDA13 mitotic detection datasets. 
Candidate nuclei are identified from blue-ratio images of stain normalized H\&E images for both the datasets. 
For ICPR12, 25 images and 10 images are used for training and validation respectively. 
Similarly for AMIDA13, 230 images and 81 images at patient level are used as training and validation respectively. 
We randomly divide both training sets into subsets to evaluate our framework for different percentage of training data.
The image and patch distributions for both datasets are provided in Table \ref{table:icpr12} and \ref{table:amida13}.

\subsection{Network parameters and training}
We employ a basic-wide WRN~\cite{zagoruyko16} (with 40 layers depth, 16 base filters, 2 widening factor and no average pooling) as the feature extractor. A 1$\times$1 convolution, global average pooling, and L2-normalization are used to produce an embedding into 128 dimensions. 
A softmax function provides the final class label. Batch normalization and ReLu activation precede all the convolution operations. 
Network weights are initialized with He normalization and $\lambda$ coefficient is set to $10^{-4}$. 
Data augmentation is performed during training using mirroring, translation and rotation.
Training begins with selecting 32 patches per batch of both mitotic and non-mitotic patches. Hard-negative non-mitotic patches are included after 20 and 40 epochs. Adam optimizer is used with learning rates set to $10^{-3}$, $10^{-3}$ and $10^{-4}$, respectively, at 0, 20 and 40th epoch. Learning rate is reduced by 0.5 when validation F1 does not improve for 5 consecutive epochs. Early stopping with a patience of 20 epochs is used to prevent network overfitting. 
The number of non-mitotic patches is later increased in proportion to the hard negative samples added. 
Coefficients $\alpha$ and $m$ for triplet loss are selected from [0.01, 0.1, 0.5] and [0.5, 1, 1.5], respectively,  and $\beta$ in DWS is set to 0.1. The model with the best validation F1 score for mitotic class is selected as the final model. At test time, for each nuclei, we predict class probabilities for 5 patches within 10 pixel radius, and aggregate their predictions to provide the final class label.

\begin{table}[t]
\caption{Mitotic detection F1 scores for ICPR 2012.}
\label{table:icpr12}
\centering
\begin{tabular}{l c c c c c}
\hline
\%Training data     & 20 & 40 & 60 & 80 & 100 \\ 
\hline
\#images   & 5      & 10     & 15   & 20  & 25 \\ 
\#mitotic & 38  & 63 & 93    & 137   & 171 \\ 
\#non-mitotic & 3.5K & 63K & 93K & 137K & 171K \\ \hline
IDSIA~\cite{ciresan13} & - & - & - & - & 0.782 \\
SegMitos~\cite{cli19} & - & - & - & - & 0.771 \\
DeepMitosis~\cite{cli18} & - & - & - & - & \textbf{0.832} \\ \hline
BCE        & 0.712    & 0.724    & 0.766    & 0.789    & 0.806 \\ 
BCE+Tr-RS  & 0.685    & 0.743    & 0.770    & 0.783    & 0.805 \\ 
BCE+Tr-BHS & 0.732    & 0.756    & 0.781    & 0.797    & 0.806 \\ 
BCE+Tr-DWS & 0.735    & 0.766    & 0.781    & 0.798    &\textbf{ 0.812} \\ 
\hline
\end{tabular}
\end{table}

\begin{table}[t]
\caption{Mitotic detection F1 scores for AMIDA 2013.}
\label{table:amida13}
\centering
\begin{tabular}{lcccccc}
\hline
\%Training data & 5 & 10 & 25 & 50 & 75 & 100 \\ 
\hline
\#images      & 12  & 24  & 61     & 120   & 176 & 230  \\ 
\#mitotic     & 28  & 48  & 107    & 228   & 340 & 452  \\ 
\#non-mitotic & 6K  & 13K & 32K    & 63K   & 92K & 120K \\ \hline
IDSIA~\cite{ciresan13} & - & - & - & - & - & 0.611 \\
SegMitos~\cite{cli19}  & - & - & - & - & - & \textbf{0.673} \\ \hline
BCE        & 0.371    & 0.443    & 0.511    & 0.577    & 0.618 & 0.644 \\ 
BCE+Tr-RS  & 0.407    & 0.467    & 0.554    & 0.589    & 0.623 & 0.641 \\ 
BCE+Tr-BHS & 0.407    & 0.492    & 0.569    & 0.616    & 0.643 & 0.652 \\ 
BCE+Tr-DWS & 0.426    & 0.531    & 0.576    & 0.627    & 0.649 & \textbf{0.669} \\ 
\hline
\end{tabular}
\end{table}

\begin{figure}[t]
\centering
\includegraphics[width=0.65\columnwidth]{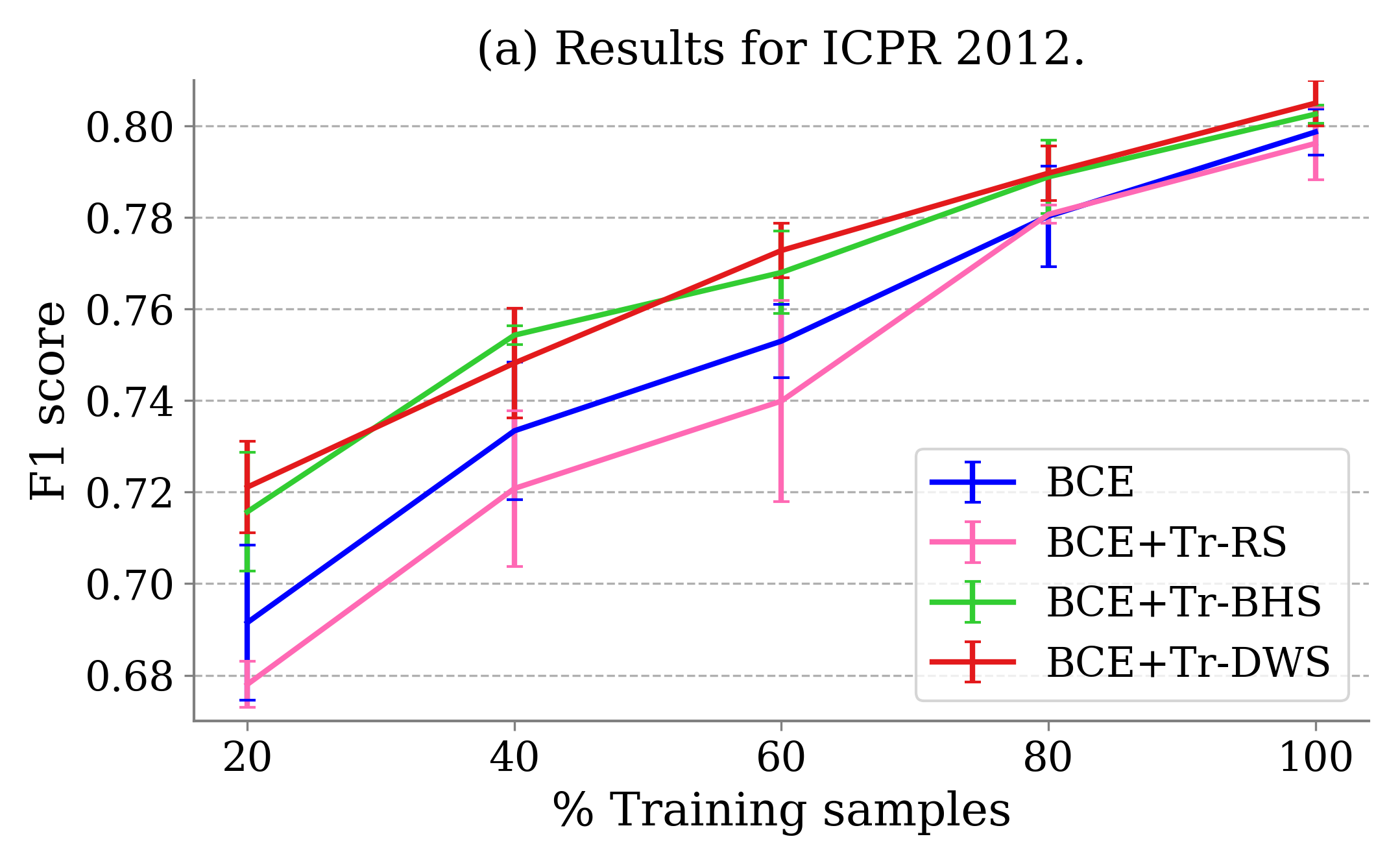}

\includegraphics[width=0.65\columnwidth]{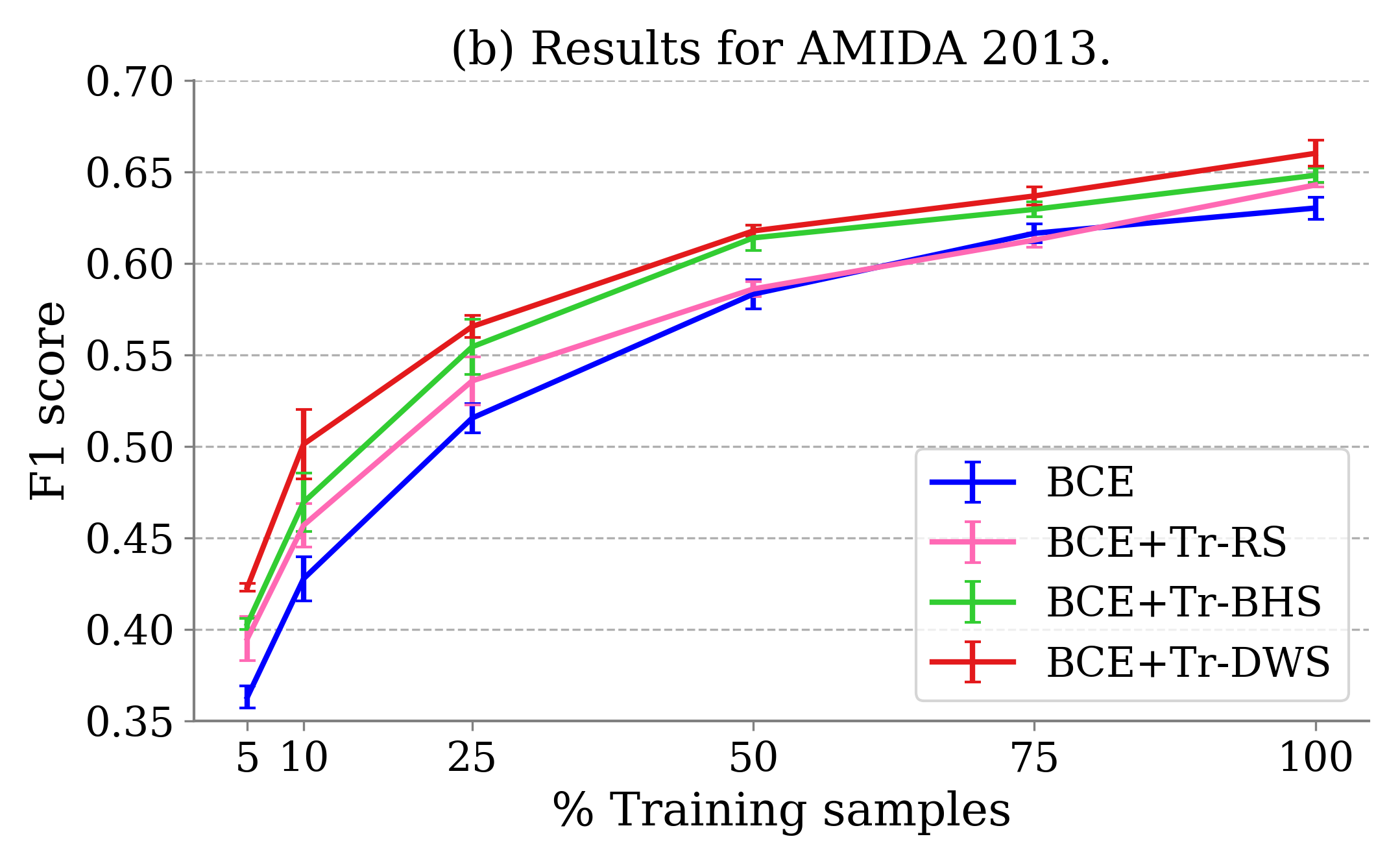}
\caption{F1 scores across trained models with different percentage of training data for (a) ICPR12 and (b) AMIDA13.}
\label{fig:results}
\end{figure}

\subsection{Network evaluation and analysis}
Mitosis detection performance is measured as the overall F1-Score, where ground truth mitoses on all test images of all patients are considered a single data set. A comparison of networks trained with different percentages of training data using BCE loss and joint loss, denoted by BCE+Tr, for different triplet sampling strategies on ICPR12 and AMIDA13 are presented in Table \ref{table:icpr12} and Table \ref{table:amida13}. 
The F1-scores for SOTA classification approaches, IDSIA and DeepMitosis, and semantic segmentation approach, SegMitos, are presented for comparison. 
For each experiment we train three networks with different initializations, and report mean and standard deviation of F1-scores for the trained networks in Figure \ref{fig:results}.

The results indicate convincing performance gains by the joint learning strategy over pure classification scheme. The improvement is significant when the training data is small, thereby confirming the benefit of the additional local information. 
It is observed that better performance is achieved for higher $\alpha$ when the training set is small, and better performance is achieved for a lower $\alpha$ as the training data increases. 
We can reason that the information from class labels is limited for small training data, therefore a strong regularization of the embeddings utilizing the local information is boosting the performance. The results also indicate the importance of sampling informative triplets. 
BCE+Tr-BHS and BCE+Tr-DWS provide consistent improvement compared to BCE. 
BHS may sometimes provide too hard negatives, causing the gradient to have high variance, whereas, DWS provides a wide range of samples, thus steadily providing informative samples while controlling the variance. 
We achieve SOTA for the classification approach and comparable performance to SOTA for the segmentation approach on both the datasets for using entire training data. 
Our framework takes approximately 3 seconds to process an HPF of 2084$\times$2084 pixels from ICPR12 on a single node of Tesla V100 GPU, which is comparable to the execution time of DeepMitosis and SegMitos. 
Thus, our framework provides the benefits of classification approach for detection with improved performance and fast execution.

\section{Conclusion}
\label{sec:conclusion}

In this paper, we present a deep learning based mitosis detection framework that simultaneously employs classification and deep metric learning to enhance the detection performance. The framework utilizes local information to regularize the feature distribution and alleviates overfitting when the training data is small. We also emphasize on the selection of samples during the training phase, and propose to use a distance-weighted sampling strategy to provide steady informative samples. Extensive
experiments are conducted on two benchmark datasets, and the results demonstrate the effectiveness of our proposed method. In the future, we aim to further exploit the sample distribution in the embedding space to improve the performance, and extend our framework to address similar tasks in digital pathology suffering from data scarcity issues, such as nuclei characterization, tumor stratification, tumor grading, etc.

\end{document}